\documentclass[conference]{IEEEtran}
\IEEEoverridecommandlockouts
\usepackage{cite}
\usepackage{amsmath,amssymb,amsfonts}
\usepackage{graphicx}
\usepackage{textcomp}
\usepackage{xcolor}

\usepackage{hyperref} 

\usepackage{multicol}
\usepackage{amsfonts,amsthm, verbatim,subcaption,mathtools}
\usepackage{paralist}
\usepackage{float,bbold}
\usepackage{dsfont}
\usepackage{algpseudocode}
\usepackage{algorithm}
\usepackage{todonotes}
\usepackage{newfloat}
\usepackage{listings}
\def\BibTeX{{\rm B\kern-.05em{\sc i\kern-.025em b}\kern-.08em
    T\kern-.1667em\lower.7ex\hbox{E}\kern-.125emX}}
    
\begin{document}

\title{Counterfactual Explanation for Regression via Disentanglement in Latent Space}

\author{\IEEEauthorblockN{Xuan Zhao}
\IEEEauthorblockA{\textit{Innovation \& digitale Transformation} \\
\textit{SCHUFA Holding AG}\\
Germany \\
xuan.zhao@schufa.de}
\and
\IEEEauthorblockN{Klaus Broelemann}
\IEEEauthorblockA{\textit{Innovation \& digitale Transformation} \\
\textit{SCHUFA Holding AG}\\
Germany \\
Klaus.Broelemann@schufa.de}

\and
\IEEEauthorblockN{Gjergji Kasneci}
\IEEEauthorblockA{\textit{Responsible Data Science} \\
\textit{Technical University of Munich}\\
Germany \\
gjergji.kasneci@tum.de}
}

\maketitle

\begin{abstract}
Counterfactual Explanations (CEs) help address the question: How can the factors that influence the prediction of a predictive model be changed to achieve a more favorable outcome from a user's perspective? Thus, they bear the potential to guide the user's interaction with AI systems since they represent easy-to-understand explanations. To be applicable, CEs need to be realistic and actionable. In the literature, various methods have been proposed to generate CEs. However, the majority of research on CEs focuses on classification problems where questions like ``What should I do to get my rejected loan approved?" are raised. In practice, answering questions like ``What should I do to increase my salary?" are of a more regressive nature. In this paper, we introduce a novel method to generate CEs for a pre-trained regressor by first disentangling the label-relevant from the label-irrelevant dimensions in the latent space. CEs are then generated by combining the label-irrelevant dimensions and the predefined output. The intuition behind this approach is that the ideal counterfactual search should focus on the label-irrelevant characteristics of the input and suggest changes toward target-relevant characteristics. Searching in the latent space could help achieve this goal. We show that our method maintains the characteristics of the query sample during the counterfactual search. In various experiments, we demonstrate that the proposed method is competitive based on different quality measures on image and tabular datasets in regression problem settings. It efficiently returns results closer to the original data manifold compared to three state-of-the-art methods, which is essential for realistic high-dimensional machine learning applications. Our code will be made available as an open-source package upon the publication of this work.
\end{abstract}


\section{Introduction}
\label{sec:intro} Machine learning models have recently become ubiquitous in our society, and the demand for explainability is on the rise. Especially in high-stakes applications like healthcare, finance, employment, etc., explainability of the model's behavior is a prime prerequisite for establishing trust in machine learning-based decision systems. Researchers have developed various techniques to explain the relationships between the input and output of a model. While models with relatively simple designs (e.g., logistic regression, decision trees, rule fit algorithm) can be interpreted straightforwardly, more complex models could be analyzed based on simpler surrogate models. These surrogate models imitate the more complex model in a post-hoc style locally and provide explanations in terms of local feature importance (i.e., for a given input). 
Another important line of research aims to answer the following question in a classification setting: ``How can the input be changed to achieve a prediction representing the favored class instead of the unfavored one?". The exploration of outcomes in alternative, similar yet non-occurring worlds is called counterfactual analysis \cite{menzies2020}. With a pre-trained and fixed model, the only way for the model to produce a different output is by altering the input. To this end, counterfactual explanations (CEs) provide prescriptive suggestions on the features of the query sample (i.e., input) that have to change (and also by how much they need to change) to achieve the desired outcome. We require this change to be minimal (i.e., associated with low costs) and actionable (i.e., realistic and feasible). With a clear semantic implication and a common logical grounding, CEs are generally easy to understand by end-users \cite{fernandez2020}. In fact, in the literature, the most direct potential use of CEs is to provide advice and guidance to end-users.

To generate practical CEs, various requirements are proposed in the literature: CEs should be actionable, sparse, valid, proximate, and computationally efficient to generate~\cite{verma2020}. Different methods have been proposed that aim to meet these requirements. It is also clear that there might be a trade-off between these requirements. However, all methods face the following problems: 
\begin{itemize}
\item The high computational costs of the constrained search for adequate CEs, especially in the case of high-dimensional input space. 
\item The generation of out-of-sample CEs, which leads to suggestions that are not actionable (i.e., unlikely or even impossible to be achieved since they do not correspond to the training data distribution).
\item Uncertainties of the user about the implementation of the suggested CEs.
\end{itemize}

\begin{figure}
\centering
\includegraphics[width=0.45\textwidth]{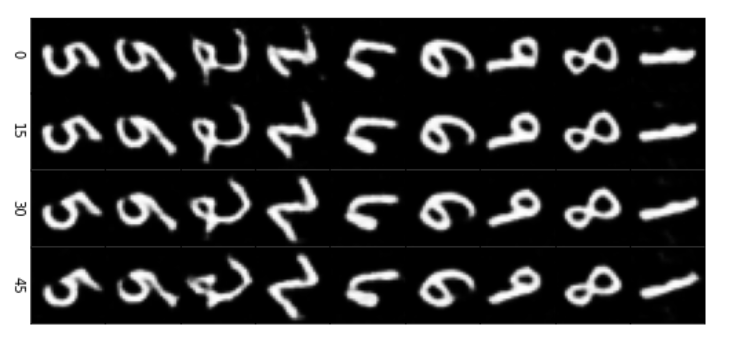}
\caption{Gradual changes of digits from rotational 0 to 45 degrees (We consider a regression task of
estimating rotation angles of MINIST images), as expected, show that the label-irrelevant characteristics like writing styles do not change during the search.}
\label{fig:mnist}
\end{figure}

Nonetheless, it is well understood that an actionable CE should be close to the data manifold and suggest meaningful (i.e., realistic and easily achievable) changes to a query sample. While most of the existing work on CEs focuses on classification models, in many practical scenarios, it is quite essential to answer questions in a regressive way, e.g., ``what should I do to improve my credit score?" Since the model outputs are discrete for classification, the notion of ``change" for a model's output is unambiguous -- when the decision boundary is crossed. In contrast, regression models have continuous outputs which can change under arbitrarily small perturbations \cite{spooner2021}. In this situation, the existing search algorithms designed for classification are unsuitable for regression problem settings. A naive way to transform regression to classification is simply binning the target labels. However, such a transformation would not address the counterfactual search within the binned intervals. To address the problems mentioned above, we develop a novel method to generate CEs for a pre-trained regressor by first disentangling the label-relevant from the label-irrelevant dimensions with the help of a special adversarial design (see details in Section \ref{sec:method}) in the latent space. CEs are then generated by combining the label-irrelevant dimensions and the predefined output. 
We show that our method maintains the characteristics of the query sample during the counterfactual search. In the problem setting of Figure \ref{fig:mnist}, we assume the counterfactual question is ``what should be changed if we want the original handwritten digit images with 0 rotational degree to become images with 45 rotational degree?" According to our method, the label-irrelevant characteristics like writing styles remain intact when the rotational angles (label) change during the search. Our method only requires access to the predictions of a previously trained regressor that we aim to explain and its training dataset. We demonstrate that it is computationally less expensive and more robust than other state-of-the-art methods in Section \ref{sec:eva}. Our main contributions are: (1) a model-agnostic framework for finding actionable CEs that lie on the data manifold with low computational expense for regression problem settings, (2) a novel strategy for manipulating the latent space to enable efficient counterfactual search, (3) a comparison of methods across image and tabular datasets.

The remainder of this paper is structured as follows: First, we briefly review the existing methods of counterfactual explanation in Section \ref{sec:back}. Then, in Section \ref{sec:method}, we propose an adversarial disentanglement approach for computing CEs. In Section \ref{sec:eva}, we present the evaluation results of our method against three state-of-the-art methods. Finally, Section \ref{sec:con} concludes the paper.

\section{Background and Related Work}\label{sec:back}
CE methods fall into two main categories: model-agnostic or model-specific. The model-specific methods usually require full access to the specific model, which needs to be explained, in order to exploit its internal structure. On the other hand, model-agnostic methods work for arbitrary machine learning (ML) models and often only require access to the predictions of an already fitted model that needs to be explained. In this paper, we focus on the model-agnostic situation. We will present some relevant work below.

\subsection{Generating Counterfactuals by Perturbing the Original Input Space}

One large branch of the literature generates counterfactuals by perturbing the original input feature space. Inverse classification \cite{lash2017} maintains sparsity by partitioning the features into immutable and mutable features and imposing budgetary constraints on the mutable features. A sampling approach is proposed by Laugel et al. \cite{laugel2017} using the growing spheres method to traverse the input space for CEs. Gradient descent is utilized in the input space to find contrastive explanations \cite{dhurandhar2018}, which are separated into pertinent positives and negatives. The authors include an autoencoder loss to keep the explanations within the data manifold. Gradient descent methods are improved with the introduction of prototypes, guiding the gradient descent towards the average value of the target class by averaging the representations of the training set in the latent space \cite{vanlooveren2020}. CE generation with reinforcement learning \cite{samoilescuModelagnosticScalableCounterfactual2021}, on the other hand, proposes a fast, model-agnostic strategy by replacing the usual optimization procedure with a learnable process. Moreover, GRACE \cite{le2020} is designed for neural networks on tabular data and combines the concepts of contrastive explanations with interventions by performing constrained gradient descent and adding a loss that measures information gain to keep the resulting explanations sparse. Many of the methods mentioned above already include generative models to obtain CEs on the data manifold. However, perturbing inputs without proper regularization can generate unconvincing and infeasible CEs \cite{goyal2019b} that resemble adversarial samples. Practical desiderata for the generated CEs, such as actionability, sparsity, validity, proximity, and computational efficiency, are included in the methods as constraints of the optimization design, which might lead to complicated optimization, especially when the input is high-dimensional.

\subsection{Generating Counterfactuals by Perturbing the Latent Space}

Latent space perturbation methods utilize generative and probabilistic models in algorithm design to ensure that CEs have a high probability under the data distribution $p(X)$. StylEx \cite{lang2021} includes the classifier in the generative model and manipulates the latent space to visualize the counterfactual search. ExplainGAN \cite{samangouei2018} is a method for finding CEs for images by training multiple autoencoders and using the signal from the classifier and discriminators to inform the learned representations. Growing sphere search is performed in the latent space of a conditional variational autoencoder to generate counterfactuals \cite{pawelczyk2020}. Along similar lines, Balasubramanian et al. \cite{balasubramanian2021} employ gradient descent in the latent space of a variational autoencoder with regularization terms. Sharpshooter \cite{barr2021} searches for counterfactuals by linear interpolation in the latent space with aid from two separate autoencoders. Xspells \cite{lampridis2022} searches through the latent space with the help of a latent decision tree for CEs for short text classification. However, searching in the latent space does not guarantee sparsity in the input space, as we show in Section \ref{sec:eva}.



\begin{figure*}[bt]
    \centering  \includegraphics[width=0.8\textwidth]{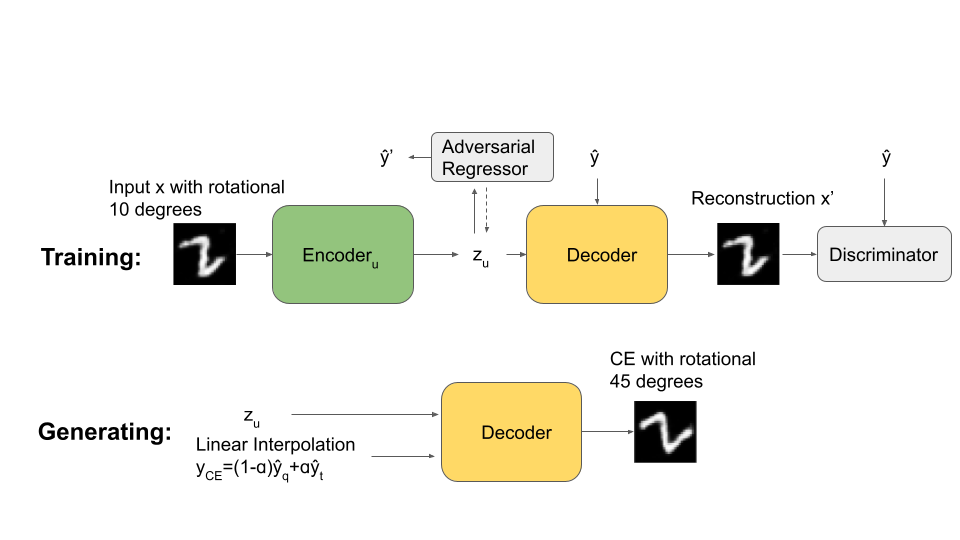}
    \caption{Diagram of the data flow in the algorithm. 
    During Training: an input sample $x$, which receives a 
    regression prediction $\hat{y}$ from the pre-trained regressor $f$. The sample $x$ is embedded into a latent space learned by the Encoder\textsubscript{u} with an Adversarial Regressor that helps remove the information related to the regression label $\hat{y}$. A Discriminator aims to discriminate the real sample $x$ from the reconstructed $x'$ by the Decoder. 
    During Generating: with a query sample $x_q$ with a regression label $\hat{y}_q$ (e.g., a rotation of 10 degrees), the label-irrelevant embedding $z_u$ is produced by running $x_q$ through the fixed Encoder\textsubscript{u} from the Training Step. Then a linear interpolation is performed by combining the predefined output $\hat{y}_t$ (e.g., an output of 45 degrees) and $\hat{y}_q$ to get $y_{CE}$. $y_{CE}$ and $z_u$ are sent through the fixed Decoder from the Training Step to get a potential CE. The potential CE goes through the pre-trained regressor $f$ to get a regression score. This process stops until the predefined output $\hat{y}_t$ is reached. 
    }
    \label{fig:diag}
\end{figure*}

\section{An Adversarial Disentanglement Approach}\label{sec:method}

\subsection{Problem Setting}
In this paper, we consider generating CEs in a regression setting. We are given a query sample (i.e., a designated input instance) for which we seek an actionable counterfactual explanation toward a predefined output. 

Formally, the training dataset of the pre-trained regressor $\hat{y_i} = f(x_i)$ is given by $\mathcal{D} = (x_i,y_i)_{i=1}^K$, where $y_i\in \mathbb{R}$ is a continuous label and $\hat{y_i}$ is the regression output/prediction. $x_q$ is a query sample with an output $\hat{y}_q$ from the regressor $f$. $\hat{y}_t$ is the predefined output. Hence, CEs are needed for the query sample $x_q$ under the regressor $f$. We consider the CEs for regression defined as below.

\textbf{Definition 1} During the CE search, if the regression output just reaches the \emph{predefined output} $\hat{y}_t$, $X_{CE}$ is the CE for the regressor $f$ with a specific predefined output $\hat{y}_t$: $f(X_{CE}) = \hat{y}_t$.

\subsection{Desiderata for CE Search}
\label{Desiderata}

Our approach is driven by the following main desiderata:

\textbf{Desideratum 1}
Intrinsic and label-irrelevant characteristics of the input query should be retained during the search process, i.e., the favoured label should be reached based on intrinsic properties of the input and with low effort.

\textbf{Desideratum 2} The generation step, should return highly realistic, i.e., feasible/actionable CEs. 

\textbf{Desideratum 3} The CEs should be retrieved efficiently also in high-dimensional settings to ensure the practical application to real-life settings.

\subsection{Learning Components}
  
We design a CE generation method based on two main steps: a Training Step and a Generating Step. It only requires access to the training dataset and prediction of the pre-trained regressor $f$ that we aim to explain through appropriate CEs. For our Training Step, we form a new training dataset $\mathcal{D}_t = (x_i,\hat{y}_i)_{i=1}^K$ by replacing original $y$ in $\mathcal{D}$ with $\hat{y}$. We then train (1) an Encoder\textsubscript{u} that captures label-irrelevant embedding $z_u$ of the original input $x$, (2) an Adversarial Regressor helps remove label-relevant information from the embedding $z_u$, (3) a Decoder that takes $z_u$  and $\hat{y}$ to output reconstruction $x'$ and (4) a Discriminator aims to discriminate the original input $x$ from the reconstruction $x'$ (see Figure~\ref{fig:diag}). In the Generating Step, we combine the label-irrelevant dimensions $z_u$ and the predefined output $\hat{y}_t$ and project them back to the original space to generate a relatively less expensive and more realistic CE. The algorithm flow is shown in Figure \ref{fig:diag}. In the following part of this subsection, we will describe each component of the Training Step in Figure \ref{fig:diag}. The details of Generating Step are described in Section \ref{sec:algorithm}.


\textbf{Encoder\textsubscript{u}} We notice that generative models usually try to generate various unseen samples. However, the generation of CEs needs to satisfy different criteria, as mentioned in Section \ref{sec:intro}. In our problem setting, to satisfy \emph{Desideratum 1} which requires maintaining the characteristics of the query sample during the search, it is intuitive to adopt disentanglement methods in the latent space of an autoencoder.


\textbf{Adversarial Regressor} Inspired by a Two-Step Disentanglement Method \cite{hadad2020}, we introduce an Adversarial Regressor to ensure that the embedding $z_u$ captured by the Encoder\textsubscript{u} is regression label-irrelevant. The motivation for us to utilize this component is that a CE should maintain the query sample's characteristics according to \emph{Desideratum 1}. It is inspired by GANs, where the discriminator gradually loses the capacity to tell the generated data from the real data during the training phase. 
While GANs are usually used to improve the quality of generated output -- telling fake from real, the adversarial component encourages the Encoder\textsubscript{u} to dismiss information about the labels, leading to disentanglement. With the Adversarial Regressor, we could guarantee that the $z_u$ and $\hat{y}$ are disentangled and independent, which prepares for the interpolation in the Generating Step. $\hat{y}'$ is the regression label of $z_u$ through the Adversarial Regressor. The \textbf{adversarial regression loss} is shown in Equation \ref{eq:reg}.

\begin{equation} \label{eq:reg}
\mathcal{L}_{Adv}=\frac{1}{N}\sum_{i=1}^{N}||\hat{y}_i-\hat{y}'_i||_2^2
\end{equation}

 \textbf{Decoder} As indicated in Figure \ref{fig:diag}, we can see that the Decoder takes regression label $\hat{y}$ and the embedding $z_u$ as inputs. In the Training Step, with the training going on, we aim to reconstruct a close sample to the query sample. We show in Figure \ref{fig:sub1_label} how the label is added to the architecture of the Decoder. $x'_i$ is the reconstruction after the Decoder. The \textbf{reconstruction loss} of the Encoder\textsubscript{u} and Decoder is shown in Equation \ref{eq:rec}.
 
 \begin{equation} \label{eq:rec}
\mathcal{L}_{Rec}=\frac{1}{N}\sum_{i=1}^{N}||x_i-x'_i||_2^2
\end{equation}

\textbf{Discriminator} To guarantee the reconstruction $x'$ is realistic and close the training data distribution to obtain actionable CEs (\emph{Desideratum 2}), we borrow the ideas of GANs again. 
Existing conditional GANs (cGANs) are mainly used for class labels, and the discriminator's loss function is shown in Equation \ref{eq:cga1a}.
Inspired by the CcGAN architecture introduced in recent work~\cite{dingContinuousConditionalGenerative2021}, we aim to use the discriminator in CcGAN to assure the quality of reconstruction $x'$. In CcGAN, a reformulation of the loss function of cGAN based on kernel density estimates is introduced, which leads to two improved empirical discriminator losses for regression labels based on the cGAN for classification -- the hard vicinal discriminator loss (HVDL) and the soft vicinal discriminator loss (SVDL). After checking the reported performance in the paper of CcGAN, we find that the performance between the two loss functions is similar.  

Hence, we adopt HVDL by \textbf{plugging} $\hat{p}^{HVE}_r(x,y)$ and $\hat{p}^{HVE}_g(x,y)$ in Equations \ref{eq:cgan2} and \ref{eq:cgan3} back to Equation \ref{eq:cga1a} to get a \textbf{discrimination loss} $\mathcal{L}_D$. We also \textbf{replace} all labels $y$ in Equations \ref{eq:cga1a}, \ref{eq:cgan2} and \ref{eq:cgan3} by regression labels $\hat{y}$ (e.g., $y_i^r$ to $\hat{y}_i^r$ and $y_i^g$ to $\hat{y}_i^g$) because we perform the Training Step on the new training dataset $\mathcal{D}_t = (x_i,\hat{y}_i)_{i=1}^K$ instead of the original dataset $\mathcal{D}$. How the regression label is input into the network of the discriminator is shown in Figure \ref{fig:sub2_label}.

$x_i^r$ and $x_i^g$ are the original/real and the reconstructed/fake $i'$th sample; $y_i^r$ and $y_i^g$ are the corresponding regression labels of $x_i^r$ and $x_i^g$; $k$ and $\sigma$ are two positive hyper-parameters, $C_1$ and $C_2$ are two normalizing factors ensuring $\hat{p}^{HVE}_r$ and $\hat{p}^{HVE}_g$ valid probability density functions; $N^r_{y,k}$ is the number of the labels $y_i^r$ satisfying $\lvert y-y^r_i\lvert \leqslant{k}$; $N^g_{y,k}$ is the number of the labels $y_i^g$ satisfying $\lvert y-y^g_i\lvert \leqslant{k}$. 
The terms in the first square brackets of $\hat{p}^{HVE}_r$ and $\hat{p}^{HVE}_g$ show that we estimate the marginal label distributions ($p_r(y)$ and $p_g(y)$) through kernel density estimates. The terms in the second square brackets are based on the assumption that a small perturbation around $y$ leads to negligible changes to $p_r(x|y)$ and $p_g(x|y)$. 

\begin{align} 
\phantom{i+j+k}
   &\begin{aligned}
\mathllap{\mathcal{L}_{D}}&=  -\mathds{E}_{y\sim p_{r}(y)}[\mathds{E}_{x\sim p_{r}({x|y})}[\textrm{log}D(x,y)]]\\ &\quad-\mathds{E}_{y\sim p_{g}(y)}[\mathds{E}_{x\sim p_{g}({x|y})}[1-\textrm{log}D(x,y)]] \\
&=-\int{\textrm{log}(D(x, y))p_{r} (x, y)dxdy}\\&\quad-\int{\textrm{log}(1-D(x, y))p_{g} (x, y)dxdy}\label{eq:cga1a}
  \end{aligned}\\
  &\begin{aligned} 
\mathllap{\hat{p}^{HVE}_r(x,y)}
&=C_1\left[\dfrac{1}{N^r}\sum_{j=1}^{N^r}\textrm{exp}\left(-\dfrac{(y-y_j^r)^2}{2\sigma^2}\right)\right] \\&\quad
\left[\dfrac{1}{N^r_{y,k}}\sum_{i=1}^{N^r}\mathbb{1}_{\{\lvert y-y^r_i \lvert \leqslant{k}\}}\delta(x-x^r_i)
\right]
\label{eq:cgan2}
  \end{aligned}\\
  &\begin{aligned} 
\mathllap{\hat{p}^{HVE}_g(x,y)}
&=C_2\left[\dfrac{1}{N^g}\sum_{j=1}^{N^g}\textrm{exp}\left(-\dfrac{(y-y_j^g)^2}{2\sigma^2}\right)\right]\\&\quad
\left[\dfrac{1}{N^g_{y,k}}\sum_{i=1}^{N^g}\mathbb{1}_{\{\lvert y-y^g_i\lvert \leqslant{k} \}}\delta(x-x^g_i)\right]
\label{eq:cgan3}
  \end{aligned}
\end{align}

\textbf{Summary} The configuration of the network in the Training Step is composed of three network branches: first, the Adversarial Regressor is trained to minimize the \textbf{adversarial regression loss} $\mathcal{L}_{Adv}$ in Equation \ref{eq:reg} -- it is trained to regress $z_u$ to $\hat{y}$. Second, the autoencoder network is trained to minimize the \textbf{total loss} $\mathcal{L}$, the sum of three terms as shown in Equation \ref{eq:recon}: (i) the \textbf{reconstruction loss} $\mathcal{L}_{Rec}$ as shown in Equation \ref{eq:rec}, (ii) minus the \textbf{adversarial regression loss} $\mathcal{L}_{Adv}$ in Equation \ref{eq:reg}, and (iii) minus the \textbf{discrimination loss} $\mathcal{L}_D$ in Equation \ref{eq:cga1a} (with adjustment). Third, the discriminator network is trained to discriminate the reconstructed from the real samples by minimizing the \textbf{discrimination loss} $\mathcal{L}_D$. 

\begin{equation} \label{eq:recon}
\mathcal{L} = \mathcal{L}_{Rec}-\lambda_{Adv} \mathcal{L}_{Adv}-\lambda_{D} \mathcal{L}_{D}
\end{equation}




\subsection{Algorithm} \label{sec:algorithm}
Algorithm \ref{alg:main} and \ref{alg:main2} show the pseudo-code of the Training and Generating Steps in Figure \ref{fig:diag}, which is used to generate the CEs in Section~\ref{sec:eva}.

\begin{algorithm}[bt]

\caption{Training Step of our proposed architecture}
   \begin{algorithmic}[1]

   \Require $\psi$,  $\phi$ , $\omega$ and $\sigma$ the initial parameters of Encoder\textsubscript{u}, Decoder, Adversarial Regressor and Discriminator;  $n$  the number of iterations; $\lambda_{Adv}$  the weight of regularization term $\mathcal{L}_{Adv}$
  
  \While {not converged} 
    \label{alg:main}
    \For{$i=0$ to $n$}
    
      \State Sample $\{x, \hat{y}\}$ a batch from dataset $\mathcal{D}$.
      \State $\omega \stackrel{+}{\gets} \bigtriangledown_{\omega} \mathcal{L}_{Adv}$
 
      \State $\sigma \stackrel{+}{\gets} \bigtriangledown_{\sigma} \mathcal{L}_{D}$

      \State Sample $\{x, \hat{y}\}$ a batch from dataset $\mathcal{D}$.
      \State $\mathcal{L} \gets \mathcal{L}_{Rec}-\lambda_{Adv} \mathcal{L}_{Adv}-\lambda_{D} \mathcal{L}_{D}$
      \State $\psi,\phi \stackrel{+}{\gets} \bigtriangledown_{\psi,\phi} \mathcal{L}$
     \EndFor 
  \EndWhile
   
   \end{algorithmic}
   \end{algorithm}

\begin{algorithm}[bt]

\caption{Generating Step of our proposed architecture}
   \begin{algorithmic}[1]

  \Require $S$ samples $\alpha=\alpha_{s=1}^S$ in $(0,1]$; $f$ regressor; $x_q$ the query sample; $\hat{y}_q$ regression label of the query sample; $\hat{y}_t$ predefined regression output; $tol$ tolerance; $x_s$ potential CE
  \For {$\alpha_s$ in $\alpha$}
   
    \State $z_u \gets Encoder\textsubscript{u}(x_q)$
     \State $\hat{y}_{CE} = (1- \alpha)\hat{y}_q + \alpha \hat{y}_t$
    \State $x_s \gets 
    Decoder(z_u,\hat{y}_s)$
    \If {$|f(x_s)-\hat{y}_{CE}|< tol$}  
     \State $x_{CE}=x_s$
     \EndIf
   \EndFor
   \State\Return {$x_{CE}$}
   
   \end{algorithmic}
   \label{alg:main2}
   \end{algorithm}



For our Training Step, 
we sample a batch from $\mathcal{D}_t$ to update the parameters of the Adversarial Regressor and Discriminator ($\omega$ and $\sigma$). Then we sample a batch from $\mathcal{D}_t$ to update the parameters: $\psi$ of the Encoder\textsubscript{u} and $\phi$ of the Decoder under the loss function introduced by Equation \ref{eq:recon}. We iterate this procedure until convergence of the loss functions is reached. 
At the end of the Training Step: we should get (1) an Encoder\textsubscript{u} that could extract the regression label-irrelevant dimensions $z_u$ from the query sample, (2) an Adversarial Regressor that could not tell the regression label $\hat{y}$ of a given input $x$, (3) a Decoder that takes the input of the encoded $z_u$ and the regression label $\hat{y}$ to reconstruct a realistic sample $x'$, and (4) a Discriminator which fails to tell the reconstructed samples $x'$ from the real input $x$. 

In our Generating Step, we are motivated to do a linear interpolation because it is relatively less computationally expensive than other existing perturbation methods (\emph{Desideratum 3}). We first pass a query sample $x_q$ with an unfavorable regression output $\hat{y}_q$ through the Encoder\textsubscript{u} to obtain the sample’s label irrelevant embedding $z_u$. Then we generate CEs by increasing or decreasing the query regression label $\hat{y}_q$ according to the task (increase or decrease the label toward the predefined output $\hat{y}_t$) 
using linear interpolation in latent space $y_{CE} = (1- \alpha)\hat{y}_q + \alpha \hat{y}_t$ with $\alpha$ in $(0, 1]$. $y_{CE}$ and $z_u$ are then projected back to the original space through the Decoder to get the potential CE $x_s$, and the pre-trained regressor $f$ assesses the regression score of the potential CE $x_s$. The counterfactual search stops if $x_s$ satisfies the predefined regression output (within a user-specified tolerance $tol$). The search is performed by sampling between $\hat{y}_q$ and $\hat{y}_t$ with a finite number of $\alpha$, which is quite efficient compared to other perturbing methods. 

\section{Experiments and Evaluation} \label{sec:eva}

We compare our method to three other counterfactual methods introduced in Section \ref{sec:back}: Gradient Descent Method improved with Prototypes (Prototype) \cite{vanlooveren2020}, CE Generation with Reinforcement Learning  (RL) \cite{samoilescuModelagnosticScalableCounterfactual2021} and Gradient Descent in the Latent
Space of a VAE (GDL) \cite{balasubramanian2021} on three datasets: MNIST \cite{zotero-844} (image), Car Price Prediction \cite{CarPricePrediction} (tabular), and House Sales in King County \cite{HouseSalesKing} (tabular). The reason why we choose Prototype and RL is that they are both designed not only for image datasets but also for tabular datasets. Besides, they both use autoencoders to learn the representation to remain close to the data distribution, similar to our design, while they use other constraints to ensure desiderata like sparsity. We use the package ALIBI \cite{klaiseAlibiExplainAlgorithms2021} for Prototype and RL to stay close to the original design. Since the application of Prototype and RL is for classification in ALIBI, we bin the continuous target labels into 20 classes. GDL is a relatively simple baseline, but it is similar to our method because both operate in latent space. We implement GDL from scratch since the source code is not available. For the experiments in this Section, the predefined output $\hat{y}_t$ is set to be $\hat{y}_t-\hat{y}_q=0.2$ (after normalization) for further comparison among different CE methods. Our test are run on an Intel(r) Core(TM) i7-8700 CPU. The networks in the experiments are built based on Pytorch \cite{NEURIPS2019_9015}.

\subsection{Measures for Comparison} We evaluate the quality of CEs by the measures taken from literature \cite{verma2020,barr2021,wachterCounterfactualExplanationsOpening2018a}:  (i) \textbf{counterfactual generation time}, time required to find a CE for a given query sample (ii) \textbf{validity}, the
percentage success in generating CEs that the predefined outputs by the users are reached (iii) \textbf{proximity}, distance($L_2$
norm) from query samples to CEs in original space (iv) \textbf{sparsity}, $L_1$
norm of the change vector in original space (v) \textbf{reconstruction loss} - a measure of the CE being in sample. We pass a CE through the autoencoder and measure its loss. A smaller loss indicates closer to the original data distribution because the autoencoder is trained on the same training dataset as the pre-trained regressor. Further details about the metrics used in the experiments can be found in the Appendix (\ref{sec:comparison}).

\begin{figure}[bt]
    \centering  \includegraphics[width=0.48\textwidth]{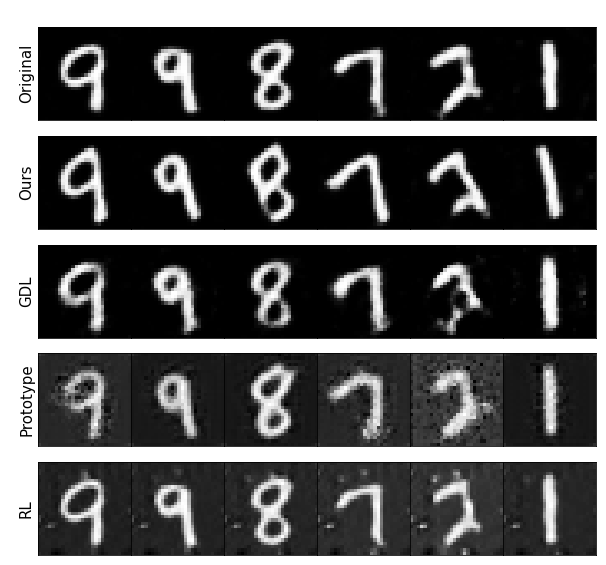}
    \caption{Original and counterfactual samples for MNIST showing from top to bottom: original query
images and CEs found via Our Method, GDL, Prototype and RL. Note Our Method exhibits more of the characteristics of the predefined output (further rotation) than the other methods. Our Method and GDL remain (perturbation in latent space) more realistic than the Prototype and RL (perturbation in original space). Additionally, only Our Method captures the higher-level concept like rotation.
    }
    \label{fig:comparison}
\end{figure}

\subsection{Datasets, Training and Evaluation}

\begin{figure*}[ht]

\subcaptionbox{The label input method for the Decoder \label{fig:sub1_label}}
  [.5\linewidth]{\includegraphics[height=5cm]{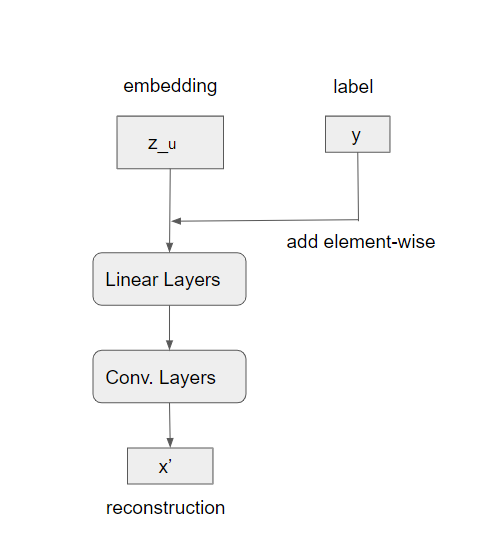}}
\subcaptionbox{The label input method for the
Discriminator\label{fig:sub2_label}} 
  [.5\linewidth]{\includegraphics[height=5cm]{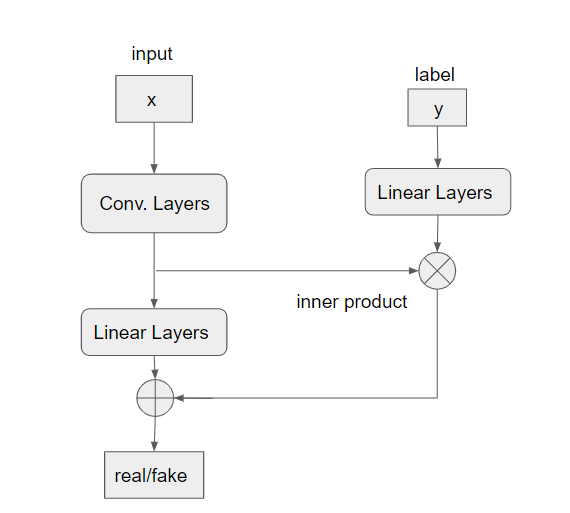}}

\caption{The label $y$ input methods of the Decoder and Discriminator for MNIST (for tabular datasets, multi-layer perceptrons are used)}
\label{fig:label}
\end{figure*}

{\bf{MNIST}} The MNIST dataset \cite{lecun-mnisthandwrittendigit-2010} is an extensive database of handwritten digits. The original MNIST dataset's target label is suitable for the multi-classification problem. Here, we appeal to its visualization and randomly rotate the original MNIST handwritten digits from 0 to 45 degrees since it is not easy to find a more proper image dataset with a regression task. We generate 20000 handwritten digit images with rotational angles randomly from 0 to 45 degrees as training and 5000 as testing sets. We adjust the problem to a regression predicting the rotational angle of the images. MNIST provides a naturally intuitive visualization of the result -- we could show a gradual change from a certain query rotational angle to the predefined output of rotational angle requested by the users. 

For MNIST, we use a CNN-based network (Figure \ref{fig:label}). We train the model for 400 epochs by stochastic gradient descent using the Adam optimizer and a batch size of 200. Our model uses two hyperparameters, $\lambda_{Adv}$ and $\lambda_{D}$, for regularizing the reconstruction loss function for the network. These weigh the effect of the adversarial losses from both the Adversarial Regressor and Discriminator. We varied $\lambda_{Adv}$ between 0.1 and 5 during training. During the training process, we want the Encoder\textsubscript{u} and Adversarial Regressor `grow' together. We found $\lambda_{Adv}= 0.5$ to give us the highest harmonic training process of the Encoder\textsubscript{u}, Adversarial Regressor and Decoder. We use this value to report our results. Changing $\lambda_{D}$ while keeping $\lambda_{Adv}$ fixed at 0.5 does not result in any significant changes, as these changes were largely overshadowed by the gains from changing $\lambda_{Adv}$. Hence, we set $\lambda_{D}= 1$ for our experiments. The details of the hyper-parameters of the Discriminator are shown in the Appendix (\ref{sec:training_minist}).

The CEs generated by our method shown in Figure \ref{fig:mnist} exhibit a combination of characteristics from the query sample (e.g., the thickness of the stroke, writing styles) and from the predefined output (e.g., a different rotational angle). Our \emph{Desideratum 1} of keeping the characteristics of the query sample which are not related to the predefined output (mentioned in Section \ref{Desiderata}) is reached. Figure \ref{fig:comparison} also shows that only our method captures the higher-level characteristics like rotation. Table \ref{tab:mnist} (mean±SD) shows the quality of counterfactual explanations measured by the above metrics. Our method outperforms the other methods in counterfactual generation time, reconstruction and validity but not always in proximity. Our Method and GDL remain (perturbation in latent space) more realistic than Prototype and RL (perturbation in original space). It also indicates that for high-dimensional applications, it is important to avoid the situation of adversarial samples mentioned in Section \ref{sec:back}, especially for methods perturbing the original space.

\begin{table*}
\centering
\caption{\label{tab:mnist}Summary of metrics for MNIST}
\resizebox{\linewidth}{!}{%
\begin{tabular}{|c|c|c|c|c|c|}
\hline
Method     & time(s)            & reconstruction           & sparsity   & validity(\%) & proximity           \\ \hline
Our Method & \textbf{0.044$\pm$0.005} &  \textbf{0.008$\pm$0.001}          & 2.343$\pm$0.627          & \textbf{90.3}   & 0.896$\pm$0.245          \\
GDL        & 5.803$\pm$0.224       & 0.520$\pm$0.135 & 0.090$\pm$0.002          & 84.2   & 2.170$\pm$0.835          \\
Prototype        & 7.345$\pm$0.784         & 4.463$\pm$0.563          & 0.014$\pm$0.005 & 75.3   & 1.075$\pm$0.235 \\
RL        & 1.804$\pm$0.112       & 5.453$\pm$0.673         & \textbf{0.012$\pm$0.008} & 65.3  & \textbf{0.875$\pm$0.132} \\
\hline
\end{tabular}}

\end{table*}

\begin{table*}
\centering
\caption{\label{tab:car}Summary of metrics for Car Price Prediction}
\resizebox{\linewidth}{!}{%
\begin{tabular}{|c|c|c|c|c|c|}
\hline
Method     & time(s)           & reconstruction           & sparsity   & validity(\%) & proximity          \\ \hline
Our Method & \textbf{0.007$\pm$0.001} & \textbf{0.012$\pm$0.003} & 2.119 $\pm$0.857        & \textbf{89.5 }    & 0.193  $\pm$0.075       \\
GDL        & 1.568 $\pm$0.047         & 0.247 $\pm$0.048        & 2.121 $\pm$0.085       & 70.9    & \textbf{0.172$\pm$0.068 }         \\
Prototype        & 1.345 $\pm$0.095         & 0.926   $\pm$0.175       & \textbf{0.014$\pm$0.005} & 58.7   & 1.075$\pm$0.055 \\
RL        & 0.085   $\pm$0.012      & 0.857$\pm$0.058         & 0.015$\pm$0.026 & 60.5  & 0.834$\pm$0.047 \\ \hline
\end{tabular}}
\end{table*}

\begin{table*}
\caption{\label{tab:hp}Summary of metrics for House Sales in King County}
\centering
\resizebox{\linewidth}{!}{%
\begin{tabular}{|c|c|c|c|c|c|}
\hline
Method     & time(s)           & reconstruction           & sparsity   & validity(\%)& proximity          \\ \hline
Our Method & \textbf{0.008$\pm$0.002} & \textbf{0.247$\pm$0.155} & 1.713 $\pm$0.973        & \textbf{91.2 }    & 3.534$\pm$0.892         \\
GDL        & 3.465    $\pm$0.765      & 0.475   $\pm$0.406      & 1.739 $\pm$0.895      & 71.2 & 1.953$\pm$0.876         \\
Prototype        & 2.164$\pm$1.973          & 0.859 $\pm$0.675         & 0.871$\pm$0.105 & 67.5   & 1.402$\pm$0.379 \\
RL        & 0.095 $\pm$0.006        & 0.870 $\pm$0.501          & \textbf{0.375$\pm$0.159} & 85.3   & \textbf{0.964$\pm$0.105} \\ \hline
\end{tabular}}
\end{table*}

{\bf{Car Price Prediction}} The Car Price Prediction dataset contains features about used cars which could be used for price prediction. For simplicity, we treat all the features as mutable. We train a regressor on the selling prices with Age of the cars, Number of Kilometres the car is driven, Number of seats as continuous features, and Fuel type of car (petrol/diesel/CNG/LPG/electric) as categorical features. 

For tabular datasets, more prepossessing is usually performed compared to image datasets. We normalize the continuous features and use one-hot encoding to deal with the categorical features. We train the model for 300 epochs by stochastic gradient descent using the Adam optimizer and a batch size of 100.  $\lambda_{Adv}$ and $\lambda_{D}$ are set at 0.5 and 0.5 for the best performance. For tabular datasets, we use a multi-layer perceptron-based network. At the end of our training step, as we expected, the Adversarial Regressor gradually outputs the mean of the prediction label $\hat{y}$ (predicted normalized car prices), which means the Encoder\textsubscript{u} extracts the label irrelevant information of the input $x$ as designed. The Adversarial Regressor could only `guess' an output. 

Based on the comparison results shown in Table \ref{tab:car}, our method outperforms in terms of time, validity, and reconstruction dimensions, while Prototype/RL is stronger in sparsity and GDL in proximity. 

{\bf{House Sales in King County}} This dataset contains house sale prices for King County, Seattle. It includes homes sold between May 2014 and May 2015. This tabular dataset contains information about the number of bedrooms, age of the house, number of bathrooms, square feet of living area, grade of the house, floor, and other features. We trained a regressor to predict the house price using five continuous and one categorical feature. From Table \ref{tab:hp}, we find that, similar to the Car Price Prediction dataset, our method is much faster at generating CEs and excels in realism, which means it can provide more actionable CEs as suggestions. The training details of this dataset are similar to the Car Price Prediction dataset, and more details are included in the Appendix.



\section{Conclusion, Limitations and Future Work}\label{sec:con}
In this work, we presented a novel model-agnostic algorithm for finding CEs via disentanglement and interpolation in the latent space for regression models. Our method implements a framework that combines label-irrelevant dimensions and the desired predefined output to generate actionable CEs for a given query sample. We demonstrated the advantages and disadvantages of our method by comparing it to three state-of-the-art methods on three different datasets (MNIST, Car Price Prediction, and House Sales in King County). The evaluation results suggest that our method provides more valid CEs that are closer to the original data distribution in a more efficient way. We also provide a visual path of how the features change while moving towards a more favorable predefined output.
Based on the presented comparison, we see further opportunities for our work to focus on improving the latent space structure to achieve better representations. Since the search is in the latent space, the sparsity of our approach is low in the original input space. However, this aligns with recent findings \cite{pawelczyk2020counterfactual} that CEs from dense regions of the data distribution tend to have higher validity and are less sparse than those from sparse regions. Future work, however, should focus on improving the sparsity of the CEs generated by generative models.

\section{Acknowledgement}
This work has received funding from the European Union’s Horizon 2020 research and innovation programme under Marie Sklodowska-Curie Actions (grant agreement number 860630) for the project “NoBIAS - Artificial Intelligence without Bias”.

\bibliographystyle{IEEEtran}  
\bibliography{references}
\appendix

\subsection{Metrics for Comparison}\label{sec:comparison}

\begin{inparaenum}[1)]
\item conterfactual generation time - time required to find a counterfactual for a query sample. Low computational expense is one of the largest strengths of our method. We report the average time spent generating counterfactual explanations for a query sample. Our approach is particularly fast compared to other methods because generating a CE requires a search through a relatively lower dimension and projection through the Decoder for no more than a fixed number of interpolated points (search stops when criteria are met). By comparison, iterative search based on gradients or perturbing in the input space can be quite expensive if the search distance is large, the learning rate is low, or the mutable dimensions are high. We do not include the time of training the Auto-encoder-based architecture with an Adversarial Regressor and a Discriminator because even though this training time is not trivial, it is a one-time cost and is not related to the scalability of CEs generation for a query sample. Ideally, our method triumphs even more, when the query sample is high dimensional. 

\item validity - the percentage of success in generating counterfactuals that indeed meet the stopping search criteria. The goal for generating CEs is that they are actually CEs, which requires them to be regressed as the target scoring. We evaluate validity by including the pre-trained regressor in the algorithm. Since linear interpolation ends when certain criteria are met for the regressor, we could see, in general, that our method performs better at this dimension. Our method uses linear interpolation to find a proper CE, while the baselines are designed with optimization for each query sample. When the optimization is designed with complicated constraints, it is intuitive that the validity is lower since valid solutions might not be available. 

\item  proximity - a measure of distance from query sample to counterfactual in original space. Proximity is usually used as a constraint to promote the performance of counterfactual generation \cite{verma2020}. This metric measures how close a counterfactual is to the query sample it is generated from. Usually this measure is calculated by the $L2$ norm in input space.

\item sparsity - a measure of features changed. Sparsity is a common metric in the counterfactual literature \cite{verma2020}. Sparsity is a measure of how sparse the change vector is. We calculate $L1$ norm of the change vector in input space). 

\item reconstruction loss - a measure of the CE being close to the data manifold. We can measure the closeness of a CE to its original dataset by passing a counterfactual through the autoencoder and measuring its reconstruction loss. During the training process of the autoencoder, we try to minimize the reconstruction loss. Intuitively, a sample with a smaller loss should be more in-sample and closer to the original data distribution. An unseen sample should have a larger reconstruction loss to the autoencoder. \end{inparaenum} 

\subsection{Further Thoughts about the Comparison Results}
We could see that our method performs better on three dimensions above: counterfactual generation time, validity and reconstruction loss due to its design. We design the architecture based on our three desiderata and in the end, through the experiment, we show that the desiderata are achieved. Our method performs linear interpolation in the latent space instead of optimization in the original space, it has the highest validity, and also because the linear interpolation happens in latent space, we cannot guarantee the sparsity and proximity in the original space, which is stated in Section \ref{sec:back}.

\subsection{Experiment Details of MNIST}\label{sec:training_minist}

The regression model used to regress the modified handwritten digits into 0 to 45 rotational angles is a CNN-based neural network with convolutional layers, fully connected layers and ReLU activations in each layer. The model is trained on standardized features with stochastic gradient descent using Adam optimizer for 100 epochs with batch
size 200. 

The Encoder, Adversarial Regressor and Decoder of our design for MNIST are also CNN-based neural networks with convolutional layers, fully connected layers and ReLU activation in each layer. The number of latent dimension is 25.  

\begin{figure}
    \centering  \includegraphics[width=0.48\textwidth]{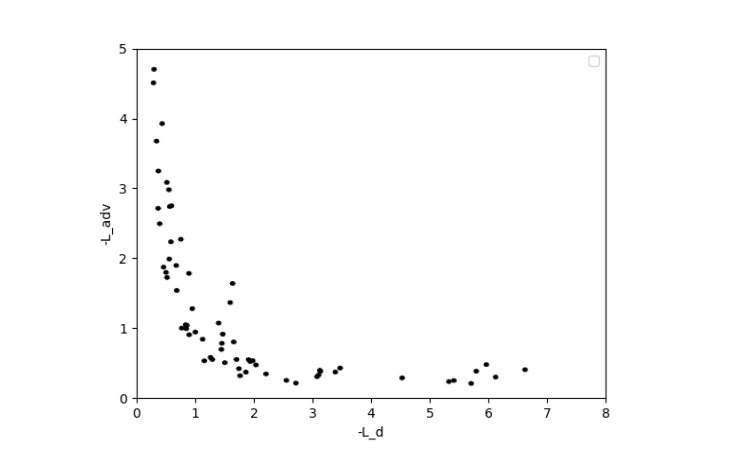}
    \caption{Pareto plot of $-L_{adv}$ versus $-L_d$ for the hyperparameter search for the MNIST dataset.}
    \label{fig:pareto}
\end{figure}

As for the hyper-parameter tuning, please refer to Figure \ref{fig:pareto} About the Discriminator, we follow the instructions from the original CcGAN paper, and set the $\sigma = 0.047$ and $k = 0.015$.

\subsection{Experiment Details of Car Price Prediction}
There is more preprocessing for tabular datasets. To stabilize the training, we remove 10 percent extreme instances (where the prices is either too high or too low). Than we normalize the continuous features and encode the categorical features with one-hot encoding. The regression model used to regress the features of the cars to their prices is is a 3 layer feedforward
neural network and ReLU activations in each layer. The model is trained on standardized features with stochastic gradient descent using Adam optimizer for 100 epochs with batch
size 100. 

The Encoder, Adversarial Regressor and Decoder of our design for Car Price Prediction are also feedforward
neural network and ReLU activations in each layer. The number of latent dimension is 2.  

As for the hyper-parameter tuning of the Discriminator, we follow the instructions from the original CcGAN paper, and set the $\sigma = 0.035$ and $k = 0.004$.

\subsection{Experiment Details of House Sales}
The preprocessing of House Sales dataset is similar to Car Price Prediction. The regression model used to regress the features of the cars to their prices is is a 3 layer feedforward neural network and ReLU activations in each layer. The model is trained on standardized features with stochastic gradient descent using Adam optimizer for 100 epochs with batch
size 100. 

The Encoder, Adversarial Regressor and Decoder of our design for Car Price Prediction are also feedforward
neural network and ReLU activations in each layer. The number of latent dimension is 2.

\end{document}